\pdfoutput=1

\documentclass[11pt]{article}

\usepackage{makecell}

\usepackage[hyphens]{url}
\usepackage{amsmath}
\usepackage[final]{acl}

\usepackage{times}
\usepackage{latexsym}

\usepackage[T1]{fontenc}

\usepackage[utf8]{inputenc}

\usepackage{microtype}
\usepackage{inconsolata}
\usepackage{array}
\usepackage{booktabs}
\usepackage{amssymb}
\usepackage{arydshln}
\usepackage{pifont}
\usepackage{tikz}

\definecolor{highlightOrange}{RGB}{255,219,170}
\definecolor{highlightBlue}{RGB}{170,222,255}
\definecolor{highlightGreen}{RGB}{219,243,183}
\definecolor{highlightRed}{RGB}{255,187,187}

\title{\textsc{GreekBarBench}: A Challenging Benchmark for Free-Text Legal Reasoning and Citations}

\author{
\\
  \textbf{Odysseas S. Chlapanis\textsuperscript{1,2}}
  \,
  \textbf{Dimitrios Galanis\textsuperscript{2,3}} 
  \,
  \textbf{Nikolaos Aletras\textsuperscript{4}} 
  \,
  \textbf{Ion Androutsopoulos\textsuperscript{1,2}}
\\
\\ 
  \textsuperscript{1}Department of Informatics, Athens University of Economics and Business, Greece \\
  \textsuperscript{2}Archimedes, Athena Research Center, Greece \\
  \textsuperscript{3}Athena Research Center, Greece \\
  \textsuperscript{4}University of Sheffield, United Kingdom\\
  \\
}

\begin{document}
\maketitle
\begin{abstract}
We introduce \textsc{GreekBarBench}, a benchmark that evaluates LLMs on legal questions across five different legal areas from the Greek Bar exams, requiring citations to statutory articles and case facts. To tackle the challenges of free-text evaluation, we propose a three-dimensional scoring system combined with an LLM-as-a-judge approach. We also develop a meta-evaluation benchmark to assess the correlation between LLM-judges and human expert evaluations, revealing that simple, span-based rubrics improve their alignment. Our extensive evaluation of $13$ proprietary and open-weight LLMs shows that even though the top models exhibit impressive performance, they remain susceptible to critical errors, most notably a failure to identify the correct statutory articles.



\end{abstract}

\section{Introduction}


As legal AI assistants become increasingly prevalent, the need for realistic legal LLM benchmarks has never been more imperative.\footnote{\url{https://www.abajournal.com/web/article/aba-tech-report-finds-that-ai-adoption-is-growing-but-some-are-hesitant}} Most widely used legal Natural Language Processing (NLP) benchmarks \citep{chalkidis-etal-2022-lexglue, niklaus-etal-2023-lextreme} focus on classification tasks, e.g., \emph{legal judgement prediction} \citep{Aletras2016PredictingJD}, which have been criticized \citep{medvedeva-mcbride-2023-legal, mahari-etal-2023-law} for being more constrained and less representative than real-world tasks. Even more recent LLM-focused legal benchmarks \citep{legalbench, fei-etal-2024-lawbench, joshi-etal-2024-il} do not go beyond closed-form questions (e.g., multiple-choice questions), failing to capture the true complexity of legal reasoning in practice, which involves identifying, analyzing and synthesizing relevant information to reach a conclusion. Unfortunately, most existing benchmarks with challenging legal questions and free-text responses are proprietary and thus inaccessible to the research community.\footnote{\url{https://www.vals.ai/benchmarks}}

\begin{table}[t]
    \centering
    \small
    \begin{tabular}{p{0.9\columnwidth}}
        \toprule
        \textbf{Facts} \\
        \midrule
        $[1]$ Antonis visited his dermatologist, Ioannis, to remove facial skin tags.\\
        $[2]$ His assistant, Penelope, took a bottle and started washing the area.\\ 
        $[3]$ The bottle contained pure acetic acid (not a solution, as prescribed), which led to burns on Antonis' face. \\
        $[4]$ He is now seeking €2.500 for treatment costs plus €75.000 for moral damages.\\
        \midrule
        \textbf{Question}\\
        \midrule
        Which individuals are liable for the injury? \\
        \midrule
        \textbf{Relevant Legal Context}\\
        \midrule
        Civil Code 914: Anyone who unlawfully causes damage must compensate the victim.\\
        Civil Code 922: An employer is liable for unlawful damages caused by their employee during work.\\
        \midrule
        \textbf{Ground Truth Answer}\\
        \midrule
        \hspace{-0.3em}\colorbox{highlightGreen}{
        Ioannis is responsible vicariously for Penelope's}\\
        \hspace{-0.3em}\colorbox{highlightGreen}{actions $[3]$}\colorbox{highlightBlue}{(\textbf{Civil Code 922})}\colorbox{highlightGreen}{and Penelope is directly}\\
        \hspace{-0.3em}\colorbox{highlightGreen}{liable for her \textbf{negligence} $[3]$}\colorbox{highlightBlue}{(\textbf{Civil Code 914}).}\\
        \hspace{-0.3em}\colorbox{highlightOrange}{\textbf{Hence, both are liable} and must compensate Antonis.}\\
        \bottomrule
    \end{tabular}
    \caption{Cropped example (English translation) from \textsc{GreekBarBench}. The answer requires multi-hop reasoning and citing legal articles and case facts. The spans corresponding to the scoring dimensions are highlighted in color: \textit{Facts} (green), \textit{Cited Articles} (blue) and \textit{Analysis} (orange). \textit{Important spans} are marked in bold and cited facts are denoted by square brackets. The complete example is presented in Appendix~\ref{app:example}.}
    \label{tab:example}
\end{table}

Another challenge is that realistic benchmarks often require costly manual evaluation by legal experts, which limits scalability \citep{hallucinationfree, martin2024bettergptcomparinglarge}. Automatic evaluation, using the LLM-as-a-judge framework \citep{llmJudge}, is a promising alternative; however, its reliability has not been extensively assessed in legal reasoning \citep{bhambhoria2024evaluatingailawbridging, li2025casegenbenchmarkmultistagelegal}.

To address these issues, we present the \textsc{GreekBarBench}, a benchmark that evaluates the reasoning capabilities of LLMs on challenging legal questions across five legal areas. The questions are taken from the Greek Bar exams and require open-ended answers with citations to statutory articles and case facts. 
In addition, we introduce an accompanying benchmark for LLM-judges, designed to measure how well their scores correlate with those of human experts.
\textsc{GreekBarBench} is the only Greek dataset for legal reasoning. Our main contributions are the following:
\begin{itemize}
\item \textbf{\textsc{GreekBarBench}}: a challenging legal reasoning benchmark that requires free-text answers citing case facts and statutory articles. 
\item \textbf{GBB-JME}: an accompanying dataset with human-evaluated answers from five different LLMs, to assess the quality of candidate LLM-judges in \textsc{GreekBarBench}.
\item A three-dimensional \textbf{scoring system} and an \textbf{LLM-judge framework} based on \textbf{span-rubrics} per dimension (\emph{Facts}, \emph{Cited Articles}, \emph{Analysis}), which aligns well with human expert evaluation.
\item A systematic \textbf{evaluation of 13 frontier and open-weight LLMs} on \textsc{GreekBarBench}, using the best LLM-judge at GBB-JME.

\item Manual error analysis to identify important weaknesses in the model responses. 
\end{itemize}

All resources including code, the two benchmarks (except for a small semi-private test set) and the prompts are publicly available.\footnote{Code and prompts: \url{https://github.com/nlpaueb/greek-bar-bench}\\Dataset: \url{https://huggingface.co/datasets/AUEB-NLP/greek-bar-bench}}

\section{\textsc{GreekBarBench (GBB)}}

\subsection{Greek Bar Exam}
\label{sec:exams}
To become licensed attorneys, Greek law graduates must pass the Greek Bar exam, which is organized by the Greek Bar Association and requires a minimum score of six out of ten. The exam evaluates candidates through practical legal questions across five key areas of law: Civil Law, Criminal Law, Commercial Law, Public Law, Lawyers' Code.

Greece operates under a statutory legal system, where laws are derived from codified statutes rather than judicial precedents. Consequently, the Greek Bar exam is an open-book test requiring candidates to locate and cite the correct statutory articles from the provided legal documents. These documents include the Civil Code and Civil Procedure Code, the Criminal Code and Criminal Procedure Code, eight Commercial Law codes, eleven Public Law codes, the Lawyers' Code, and the Code of Ethics for Legal Practice (see Table~\ref{tab:stats}).

Past Greek Bar exam papers and suggested solutions (2015–2024) are publicly available in an annually updated PDF booklet.\footnote{The booklet is available at \url{https://www.lawspot.gr/nomika-nea/panellinios-diagonismos-ypopsifion-dikigoron-themata-exetaseon-kai-endeiktikes-3}} The authors of this paper have been granted permission by the Greek Bar Association to use the questions and by the booklet's authors to use their respective solutions.

\subsection{Task}

\begin{table}[t]
    \centering
    \resizebox{\columnwidth}{!}{%
    \renewcommand{\arraystretch}{1.2}
    \begin{tabular}{| l | c | c | c | c | c |}
        \hline
        \textbf{Benchmark} & \textbf{Lang} & \textbf{Cite} & \textbf{Legal} & \textbf{Judge}\\
        & & \textbf{Articles} & \textbf{Context} & \textbf{Eval}\\
        \hline 
        LegalBench & \texttt{en} & \ding{55} & \ding{51} & \ding{55}\\
        \hline
        LexEval (5.4)& \texttt{zh} & \ding{55} & \ding{55} & \ding{55}\\
        \hline
        CaseGen & \texttt{zh} & \ding{55} & \ding{55} & \ding{55}\\
        \hline
        OAB-Bench & \texttt{por} & \ding{55} & \ding{55} & \textbf{---}\\
        \hline
        LLeQA & \texttt{fr} & \ding{51} & \ding{55} & \ding{55}\\
        \hline
        LEXam & \texttt{en, de} & \ding{51} & \ding{55} & \textbf{---}\\
        \hline
        \textbf{GBB (Ours)} & \texttt{el}& \ding{51} & \ding{51} & \ding{51}\\
        \hline
    \end{tabular}%
    }
    \caption{
    Comparison of popular legal benchmarks. `Lang': language of dataset. `Cite Articles': the legal articles must be cited. `Legal Context': the necessary legal information is provided. `Judge-Eval': separate benchmark to compare LLMs-as-judges (`--' means that the feature is not publicly available).
    }
    \label{tab:comparison}
\end{table}

The design of \textsc{GreekBarBench} is inspired by the workflow of candidate lawyers on the exam. They first study the case facts to identify legal issues, then navigate legal codes to locate the relevant chapter and pinpoint the specific statutory article to support their arguments. To mirror this process, each instance in the benchmark is derived from an exam question and consists of (1) the facts of the incident, (2) the exam question, and (3) a collection of potentially relevant chapters of statutory articles. The desired output is the free-text answer to the question, providing an \emph{analysis} with citations to the case \emph{facts} and the applicable statutory \emph{articles}. 

Table~\ref{tab:comparison} highlights the unique contributions of \textsc{GreekBarBench}. It is one of the few benchmarks to require citations to legal articles (and the only one to require citations to case facts). Furthermore, it is unique in providing a corpus of potentially relevant articles and including a dedicated framework for assessing LLMs-as-judges.

\subsection{Dataset Statistics}
We collect a total of 65 exam papers; 13 exam papers from each of the five aforementioned areas. The PDF booklet (§~\ref{sec:exams}) is converted to text format and further processed (\S~\ref{sec:fact-segmentation}) to prepare the dataset. Each exam paper includes 4.7 questions on average, resulting in a total of 310 questions. We keep the questions from 2024 (22 in total) as a semi-private test set, to avoid data contamination.\footnote{`Semi-private' means that the test set is not public, but the raw PDFs might have leaked into the pretraining data.} The public test set contains 288 questions, while the semi-private set will be updated annually with two new exam papers from each legal area.


Answering the exam questions requires citing articles from 25 legal code documents, obtained from the official website of the Greek National Printing House.\footnote{\url{https://search.et.gr/el/}} Detailed statistics for these documents are presented in Table~\ref{tab:stats} per legal area. Articles are cited 931 times in total, across all exam questions. The articles within each legal code document are grouped thematically into chapters. The total number of citable articles is more than 16,000. This extensive corpus is far too large to fit entirely within the context window of any LLM.

\begin{table}[t]
    \centering
    \resizebox{\columnwidth}{!}{%
    {\renewcommand{\arraystretch}{1.2} 
    \setlength{\tabcolsep}{2.5pt}        
    \begin{tabular}{|l|c|c|c|c|c|}
        \hline
        \textbf{Law Areas} & \textbf{Questions} & \textbf{Legal} &\textbf{Total}& \textbf{Cited} & \textbf{Context} \\
        && \textbf{Codes} &\textbf{Articles} &\textbf{Articles}&\textbf{(tokens)}\\ 
        \hline 
        Civil & 71 & 2 & 3,264 & 286  & 87k \\ \hline
        Criminal & 53 & 2 & 1,253 & 186  & 58k \\ \hline
        Commercial & 58 & 8 & 4,177 & 159  & 29k \\ \hline
        Public  & 71 & 11 & 2,912 & 118  & 67k \\ \hline
        Lawyers  & 57 & 2 & 4,476 & 182  & 66k \\ \hline
        \textbf{Total} & 310& 25& 16,082& 931  & 62k (avg)\\ \hline
    \end{tabular}
    }
    }
    \caption{Summary of dataset statistics. `Legal Codes' indicates the number of distinct legal code documents in each area. `Cited Articles' is the total number of citations to legal code articles. `Context' denotes the average token count of the relevant legal context (chapters of legal code) provided in the input of candidate LLMs.}
    \label{tab:stats}
\end{table}

\subsection{Relevant Legal Context}
\label{sec:relevant-context}

As mentioned in Section \ref{sec:exams}, the Greek Bar exams are open-book, allowing candidate lawyers to navigate legal code documents to identify relevant statutory articles for the presented case. Simulating this setup presents several challenges. One approach would be implementing a Retrieval-Augmented Generation (RAG) pipeline, using sparse (e.g., BM25) or dense retrievers \citep{karpukhin-etal-2020-dense} to select the $k$ most `relevant' articles for inclusion in the LLM's input. However, this approach suffers from three significant limitations: a) candidate lawyers taking the exams do not have access to such retrieval tools, making direct comparisons with human performance problematic; b) retrievers are prone to errors, creating a substantial risk that even with large values of $k$, the ground truth articles might not appear among the top retrieved articles; and c) as demonstrated by \citet{krishna-etal-2025-fact}, benchmarking RAG systems requires testing multiple configurations with varying values of $k$ and, ideally, different retriever models, complicating fast integration of new LLMs. 

To better simulate the exam environment, we propose an approach that mirrors how a candidate lawyer works: using chapter-level context rather than discrete retrieved articles. For each case, we first use regular expressions to extract the ground truth legal articles cited in its questions. We then identify the legal code chapters these articles belong to and aggregate every article from those chapters into a single corpus. This corpus serves as a broad legal context for all questions related to that case, more faithfully simulating how a lawyer navigates an entire chapter. The average corpus length is 62,000 tokens per question (Table~\ref{tab:stats}), which is within the capacity of most modern LLMs.

\subsection{Fact Segmentation}
\label{sec:fact-segmentation}
To streamline the evaluation process, we require citations to facts in candidate answers, though this is not mandatory in the official exams. To implement this, we segment the case facts into sentences using the Segment-Any-Text neural model \citep{frohmann-etal-2024-segment} and present them as a numbered list (as shown in Table~\ref{tab:example}). This structure makes it straightforward to detect any factual errors.


\subsection{Three-Dimensional Scoring System}
\label{sec:evaluation}
The official evaluation committee of the Greek Bar Exams grades candidate answers on a scale of 1 to 10. This process lacks explicit guidelines and relies on a holistic comparison to the ground truth, limiting its analytical depth for benchmarking LLMs.

Drawing inspiration from established legal research and evaluation practices \citep{ClarkDeSanctis2013}, and guided by our legal expert annotators (\S~\ref{sec:manual}), we develop a novel three-dimensional scoring system to improve the evaluation process for the benchmark. The proposed approach assesses legal reasoning across three dimensions: the \emph{Facts}, the \emph{Cited Articles}, and the \emph{Analysis}. Each dimension is rated on a scale of 1 to 10, and the final score is their \emph{mean}. This fine-grained framework allows the detection of specific shortcomings in the abilities of LLMs. The \emph{Facts} score measures understanding of case facts; the \emph{Cited Articles} score evaluates the ability to identify and cite applicable legal articles; and the \emph{Analysis} score evaluates the ability to construct valid legal arguments. 

\section{Automatic Evaluation}
To address the evaluation of free-text answers without the prohibitive cost of manual annotations, we use the LLM-as-a-judge framework \citep{llmJudge}. LLM-judges can be categorized into two primary types: (a) \emph{pairwise} LLM-judges, which evaluate two candidate answers and determine which is preferred (or declare a tie), and (b) \emph{grading} LLM-judges, which assign an integer score to each individual candidate answer \citep{llmJudge}. In our work, we focus on \emph{grading} LLM-judges to allow cost-effective integration of new participant LLMs without the overhead of quadratically increasing pairwise comparisons.

To improve the alignment of LLM-judges with human expert annotators, we propose novel span-based rubrics; i.e., evaluator instructions in the form of annotated spans per question. These spans will guide the LLM-Judge in what to assess in the candidate answers. However, even with these question-specific rubrics, replicating the nuanced evaluation of human experts, especially in complex tasks like legal writing, cannot be guaranteed. For this reason, we also include a framework to meta-evaluate whether LLM-judges are suitable proxies for human evaluation on \textsc{GreekBarBench}.

\subsection{Simple LLM-Judge} 
\label{sec:simple_llm_judge}

As an initial approach, we designed a straightforward prompt for a simple LLM-judge. The prompt outlines the evaluation task and explicitly defines the criteria for the \emph{Facts}, \emph{Cited Articles}, and \emph{Analysis} scores. All necessary contextual information is provided; the facts of the case, the specific legal question, the ground truth answer with the cited articles and the candidate answer to be evaluated. This context mirrors the information provided to the human annotators for the manual evaluations (\S~\ref{sec:manual}). The required output format is clearly specified: the model must provide an explanation for each score, followed by the integer score. The complete prompt is presented in Appendix~\ref{app:prompts} (Fig.~\ref{app:simple-judge-prompt}).

\subsection{Span LLM-Judge} 
\label{sec:span_llm_judge}
Inspired by \citet{ClarkDeSanctis2013}, who found that rubrics improve the consistency of legal writing evaluation, we developed a rubric-based annotation process for our benchmark. Our legal experts constructed these rubrics by first identifying reference spans in the ground-truth answer corresponding to the three scoring dimensions, marking each with a distinct color: \emph{Facts} (green), \emph{Cited Articles} (blue), and \emph{Analysis} (orange). An example of these colored spans is shown in Figure~\ref{tab:example}. Within each span, the annotators then identified \emph{important subsets}, i.e., the specific words or phrases crucial for a correct answer (shown in bold in Table~\ref{tab:example}). 

To minimize the annotation burden, we did not assign point values to these subsets, which contrasts with the approach in prior work \citep{starace2025paperbenchevaluatingaisability, pires2025automatic}. Instead, the LLM-judge is instructed to assess whether a candidate answer covers the key information within the spans, with a focus on the \emph{important subsets}, and to use this assessment to score each dimension. The full prompt is available in Appendix~\ref{app:prompts} (Fig.~\ref{app:span-judge-prompt}).

\section{Meta Evaluation}
\label{sec:meta-eval}
Meta-evaluation of \emph{grading} LLM-judges aims to quantify the alignment between LLM-generated scores and human expert annotations. Previous research has predominantly relied on Pearson's or Spearman's correlation coefficients as primary meta-metrics \citep{judgeBench, niklaus2025swiltrabenchswisslegaltranslation}, often without substantial justification. However, advancements in meta-evaluation have emerged from the machine translation domain, particularly through the WMT Metrics Shared Task \citep{freitag-etal-2024-llms, freitag-etal-2023-results}, where automatic evaluation frameworks have been systematically compared and refined. The task aims to identify optimal metrics for translation quality assessment by comparing system outputs against references. Recent findings demonstrate that state-of-the-art metrics are increasingly LLM-based. The task has revealed that Pearson's correlation coefficient exhibits vulnerability to outliers, while Spearman's $\rho$ disregards the magnitude of ranking errors, applying uniform penalties.
To address these limitations, WMT has adopted Soft Pairwise Accuracy (SPA) \citep{thompson-etal-2024-improving}, a metric that assigns partial credit for nearly correct rankings, thereby providing an evaluation framework that better reflects the alignment of metrics with human experts.

\subsection{Soft Pairwise Accuracy (SPA)}

SPA measures the degree of alignment in evaluation \emph{confidence} between human experts and LLM-judges (or any other automatic metric). For example, if a human expert is \emph{confident} that one system (e.g., a candidate LLM from \textsc{GreekBarBench}) outperforms another, but the LLM-judge is \emph{uncertain}, SPA penalizes the judge—even if the ranking happened to be correct. To do this, SPA approximates the \emph{confidence} of each judge (human or LLM) on each pairwise comparison between systems using p-values of appropriate permutation tests \citep{Fisher1935}, as detailed below. We use the original implementation.\footnote{\url{https://github.com/google-research/mt-metrics-eval}} 
Formally, SPA between a metric $m$ and human experts $h$ is defined as:

{\small
\[
\textit{SPA}(m,h) =  {\binom{N}{2}}^{-1} \sum_{i=0}^{N-1} \sum_{j=i+1}^{N-1} \left( 1 - \left| p_{ij}^h - p_{ij}^m \right| \right)
\]
}

\noindent where $N$ is the number of systems being evaluated, $p_{ij}^h$ is the p-value for the hypothesis that system $i$ is better than system $j$ according to human scores, and $p_{ij}^m$ is the corresponding p-value according to the metric under evaluation. The term ${\binom{N}{2}}^{-1}$ normalizes the summation by the total number of systems under comparison.

\paragraph{SPA permutation tests:}
To estimate \emph{confidence} of an evaluator (either human or automatic) in a pairwise system comparison, SPA uses permutation tests to calculate the expected mean difference under the null hypothesis that the systems are of equal quality. Specifically, a number of mock systems (1,000 in our experiments, following the original paper) are constructed as follows: for each question in the benchmark, the mock system is assigned either the score of system $i$ or system $j$ at random. The p-value is then computed as the proportion of mock systems for which the differences are greater than or equal to the mean difference between systems $i$ and $j$, as scored by the evaluator.

\section{Experiments}
\subsection{Models}
Our experiments evaluate a diverse range of LLMs, comprising proprietary models (from OpenAI, Google, Anthropic) and open-weight models; Deepseek-R1 \citep{deepseekai2025deepseekr1incentivizingreasoningcapability}, Gemma-3 \citep{gemmateam2025gemma3technicalreport}, and Llama-Krikri-8B \citep{roussis2025krikriadvancingopenlarge}; a model specifically pretrained for the Greek language. We accessed proprietary models and the large open-weight Deepseek-R1 through Application Programming Interfaces (APIs) provided by OpenAI, Google, and AWS. The remaining open-weight models were deployed on a cluster of eight A100 GPUs using the vLLM framework \citep{kwon2023efficient}. Due to limited resources, we performed three runs for each model. We used the default parameter configurations as specified by each model's provider.

\paragraph{Generation prompt:}
To generate responses from candidate LLMs, we designed a system and user prompt for the benchmark. The system prompt instructs the LLM to answer with citations to Greek statutory articles. The user prompt is structured to first describe the overall task, including clear instructions on the expected output format. Then it provides the numbered \emph{facts} of the case, the \emph{question} and the \emph{relevant legal context}. The original prompt templates are available in Appendix~\ref{app:prompts}.

\subsection{Manual Evaluation by Legal Experts}
\label{sec:manual}

In this section we present the manual evaluations that we collected for GBB-JME (\S~\ref{sec:jme}), our Judge Meta-Evaluation benchmark for assessing LLM-judges on \textsc{GreekBarBench}. We obtain ground truth evaluations (\emph{Facts}, \emph{Cited Articles}, \emph{Analysis} scores on a scale of 1 to 10) from two expert legal annotators who are licensed Greek lawyers with practical experience. The annotators were compensated for their time and expertise. They evaluated five LLMs on 87 questions drawn from three exam sessions (2024-A, 2023-A, and 2023-B), resulting in a total of 1,305 annotated samples. The models evaluated on all three exams were Claude-3.7-Sonnet, OpenAI-o1, GPT-4o, and Gemini-2.0-Flash. For the 2024 exam, we included the open-source Llama-3.1-70B; however, due to its poor performance and generation failures on several questions, we replaced it with Deepseek-R1 for the 2023-A and 2023-B exams. Annotations were managed with the open-source platform \textit{doccano}.\footnote{\url{https://doccano.prio.org}}

\begin{figure}[!t]
 \centering
 \includegraphics[
        width=\linewidth,
        trim=70pt 20pt 30pt 60pt,
        clip
    ]{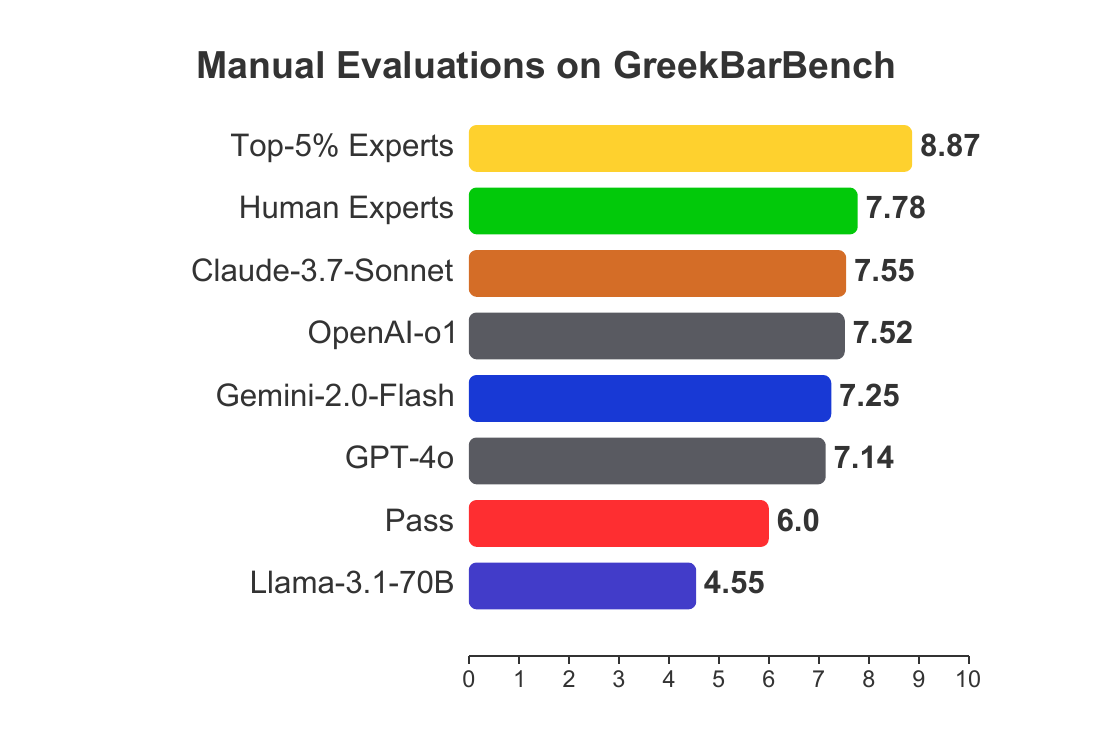}
 \caption{Manual evaluation by legal expert annotators on the semi-private test set of the 2024 exams.}
 \label{fig:manual-leaderboard}
\end{figure}
The average Krippendorff’s $\alpha$ \citep{Krippendorff2011} between the two annotators on the three-dimensional scores was 0.74, and the SPA was 0.85, both indicating a substantial level of inter-annotator agreement \citep{inter-annotator}. For the SPA calculation, we treated one annotator’s scores as ground truth and compared the other annotator’s scores against them. This differs from Section~\ref{sec:jme}, where SPA measures the correlation between LLM-generated scores and the aggregated scores of human annotators. 

As shown in Figure~\ref{fig:manual-leaderboard}, the top models in the 2024 exam were Claude-3.7-Sonnet (7.55) and OpenAI-o1 (7.52). While all LLMs except Llama-3.1-70B passed the exam (score > 6.0), they still performed below the average expert (7.78) and the top 5\% of experts (8.87).

\subsection{Judge Meta-Evaluation (GBB-JME)}
\label{sec:jme}

We evaluate seven LLMs-as-judges on our meta-evaluation benchmark, GBB-JME, using two prompts: the \emph{Simple-Judge} (\S\ref{sec:simple_llm_judge}) and the rubric-based \emph{Span-Judge} (\S\ref{sec:span_llm_judge}). To measure the consistency of the judges, we report the average and standard error over three runs.

The results, presented in Table~\ref{tab:jme}, reveal a clear trend: leading models (GPT-4.1, Gemini-2.0-Flash, and GPT-4.1-mini) generally improve with the \emph{Span-Judge} prompt while weaker ones seem to struggle with its complexity. Surprisingly, the top-performing model was GPT-4.1-mini (0.862), which outperformed its larger and theoretically more powerful variant, GPT-4.1 (0.837). The stronger performance from the smaller model as a judge corroborates findings from other work \citep{niklaus2025swiltrabenchswisslegaltranslation}. 

More generally, the results highlight a key distinction between a model's performance as a judge and as a candidate, a pattern that holds across different model families. This is illustrated by the open-weight Gemma-3-27B (0.785 with \emph{Simple-Judge}), which outperformed Gemini-2.0-Flash (0.778 with \emph{Span-Judge}) despite the latter's superior performance as a candidate (see \S\ref{sec:leaderboard}, Table~\ref{tab:leaderboard}).


Due to its strong performance and cost-effectiveness, we adopted GPT-4.1-mini with the \emph{Span-Judge} prompt for all subsequent evaluations, with the total cost amounting to approximately \$100. For those seeking a more accessible option, the open-weight Gemma-3-27B serves as a strong, low-cost alternative. To support future work, we are publicly releasing the GBB-JME benchmark, enabling other researchers to evaluate new LLMs-as-judges. This approach ensures that the benchmark's value grows over time, as it can be continuously replicated and validated with increasingly capable models serving as judges.

\begin{table}[t]
\centering
\renewcommand{\arraystretch}{1.2} 
\resizebox{\columnwidth}{!}{%
\small
\setlength{\tabcolsep}{4pt}
\begin{tabular}{l|cc|c}
\hline
\textbf{Model}           & \textbf{Simple-Judge} & \textbf{Span-Judge} & \textbf{Cost} \\
& (SPA)   & (SPA)  &\\ \hline 
GPT-4.1-mini    & \textbf{0.851} ± 0.008 & \textbf{0.862} ± 0.001 & \textbf{\$}\textbf{\$} \\ 
GPT-4.1  & 0.835 ± 0.010 & 0.837 ± 0.005& \textbf{\$}\textbf{\$}\textbf{\$} \\ 
Gemini-2.0-F     & 0.745 ± 0.009    & 0.778 ± 0.011&  \textbf{\$}\\ 
Gemma-3-27B      & 0.785 ± 0.015  & 0.733 ± 0.017 & \textbf{---} \\ 
Krikri-8B      & 0.708 ± 0.004    & 0.638 ± 0.019  & \textbf{---}   \\ 
Gemini-2.0-LF     & 0.667 ± 0.016      & 0.695 ± 0.010 & \textbf{\$}\\ 
GPT-4.1-nano    & 0.675 ± 0.006 & 0.416 ± 0.008 & \textbf{\$} \\ \hline 
\end{tabular}
}
\caption{Comparison of LLMs-as-judges on GBB-JME, using Simple-Judge and Span-Judge. The cost per one million input tokens is denoted by: {\small \$} (less than {\small \$}0.3), {\small \$}{\small \$} (less than {\small \$}1), and {\small \$}{\small \$}{\small \$} (less than {\small \$}3).}
\label{tab:jme}
\end{table}

\subsection{Results on \textsc{GreekBarBench}}
\label{sec:leaderboard}
We conduct an extensive automatic evaluation of $13$ LLMs on \textsc{GreekBarBench} (Figure~\ref{tab:leaderboard}). We use GPT-4.1-mini as the judge with the \emph{Span-Judge} prompt (\S~\ref{sec:span_llm_judge}), reporting the \emph{Facts} score, \emph{Articles} score, \emph{Analysis} score and \emph{Mean} score with standard error from three runs. The evaluation includes proprietary models such as GPT-4o, the GPT-4.1 family (GPT-4.1, GPT-4.1-mini, GPT-4.1-nano), Gemini-2.0-Flash, and Claude-3.7-Sonnet (reasoning disabled), along with the reasoning models OpenAI-o1 and Gemini-2.5-Flash. The open-weight models include the Gemma-3 family (27B, 12B, and 4B parameters), the specialized Greek model Llama-Krikri-8B-Instruct (Krikri-8B), and the reasoning model DeepSeek-R1.

\begin{table}[t]
\centering
\resizebox{\columnwidth}{!}{%
\setlength{\tabcolsep}{5pt}
\begin{tabular}{l|c@{\hspace{4pt}}c@{\hspace{4pt}}c:c}
\hline
\textbf{Model} & {\small \textbf{Facts}} & {\small \textbf{Articles}} & {\small \textbf{Analysis}} & \textbf{Mean} \\
\hline
Gemini-2.5-F & {\footnotesize 8.62} & {\footnotesize\textbf{8.16}} & {\footnotesize\textbf{8.36}} & {\footnotesize\textbf{8.38 ± 0.02}} \\
GPT-4.1 & {\footnotesize\textbf{8.65}} & {\footnotesize 8.05} & {\footnotesize 8.34} & {\footnotesize 8.35 ± 0.02} \\
OpenAI-o1 & {\footnotesize 8.24} & {\footnotesize 7.55} & {\footnotesize 7.55} & {\footnotesize 7.78 ± 0.01} \\
Claude-3.7-S. & {\footnotesize 8.34} & {\footnotesize 7.37} & {\footnotesize 7.60} & {\footnotesize 7.77 ± 0.04} \\
GPT-4.1-mini & {\footnotesize 8.28} & {\footnotesize 7.18} & {\footnotesize 7.40} & {\footnotesize 7.62 ± 0.01} \\
Gemini-2.0-F & {\footnotesize 8.28} & {\footnotesize 7.13} & {\footnotesize 7.12} & {\footnotesize 7.51 ± 0.03} \\
GPT-4o & {\footnotesize 8.15} & {\footnotesize 7.10} & {\footnotesize 7.18} & {\footnotesize 7.48 ± 0.03} \\
Deepseek-R1 & {\footnotesize 7.65} & {\footnotesize 6.48} & {\footnotesize 6.55} & {\footnotesize 6.90 ± 0.02} \\
Gemma-3-27B & {\footnotesize 7.46} & {\footnotesize \colorbox{highlightRed}{5.63}} & {\footnotesize \colorbox{highlightRed}{5.79}} & {\footnotesize 6.29 ± 0.02} \\
Krikri-8B & {\footnotesize 7.20} & {\footnotesize \colorbox{highlightRed}{5.65}} & {\footnotesize \colorbox{highlightRed}{5.74}} & {\footnotesize 6.20 ± 0.03} \\
Gemma-3-12B & {\footnotesize 7.25} & {\footnotesize \colorbox{highlightRed}{5.37}} & {\footnotesize \colorbox{highlightRed}{5.62}} & {\footnotesize 6.08 ± 0.04} \\
\hline
PASS score & {\footnotesize 6.00} & {\footnotesize 6.00} & {\footnotesize 6.00} & {\footnotesize 6.00 ± 0.00} \\
\hline
GPT-4.1-nano & {\footnotesize 6.88} & {\footnotesize \colorbox{highlightRed}{4.69}} & {\footnotesize \colorbox{highlightRed}{4.79}} & {\footnotesize \colorbox{highlightRed}{5.45 ± 0.03}} \\
Gemma-3-4B & {\footnotesize \colorbox{highlightRed}{5.53}} & {\footnotesize \colorbox{highlightRed}{3.76}} & {\footnotesize \colorbox{highlightRed}{3.95}} & {\footnotesize \colorbox{highlightRed}{4.42 ± 0.00}} \\
\hline
\end{tabular}
}
\caption{Comparison of closed and open-weight LLMs on \textsc{GreekBarBench} with GPT-4.1-mini Span-Judge on three runs. The scores reported are `Facts', `Articles', `Analysis', and `Mean' score with standard error. The best scores are shown in bold, while failed scores are highlighted in red.}
\label{tab:leaderboard}
\end{table}

Gemini-2.5-Flash (8.38) and GPT-4.1 (8.35) achieved the highest scores on \textsc{GreekBarBench} (Table~\ref{tab:leaderboard}). These scores are not directly comparable to the human expert scores (average: 7.78, top-5\%: 8.87) in Figure~\ref{fig:manual-leaderboard}, as the models were evaluated by an LLM-judge on a larger dataset (310 vs. 22 questions). The smallest models, GPT-4.1-nano (5.45) and Gemma-3-4B (4.42) are the only models that fail the exams (passing score: 6.00). The open-weight Krikri-8B model (6.20) surpasses Gemma-3-12B (6.08) and achieves performance comparable to the significantly larger Gemma-3-27B (6.29), highlighting the benefit of language-specific pretraining.

Analyzing the scoring dimensions provides valuable insights into model capabilities. A key finding is that all models except for Gemini-2.5-Flash and GPT-4.1 struggle most with the `Articles` and the `Analysis` dimensions and that's what separates them from the others. 




Table~\ref{tab:fine-comparison} presents a fine-grained comparison for six LLMs, reporting their mean score across five different legal areas (\emph{Civil}, \emph{Criminal}, \emph{Commercial}, \emph{Public}, \emph{Lawyers}) and their average (\emph{Avg}). The results show that LLMs exhibit consistent performance across all legal areas. Notably, the smaller open models, Gemma-3-27B and Krikri-8B, struggle in certain areas, failing to meet the passing grade threshold of 6.00 (indicated by red). The second-best model, GPT-4.1, matches the top performer, Gemini-2.5-Flash, in `Criminal' and `Commercial Law', but Gemini-2.5-Flash achieves slightly higher scores in the remaining three areas.

\begin{table}[!t]
 \centering
 \renewcommand{\arraystretch}{1.1} 
 \resizebox{\columnwidth}{!}{%
 \setlength{\tabcolsep}{4pt}
 \begin{tabular}{l|ccccc:c}
 \hline
  \textbf{Model} & {\textbf{Civil}} & {\textbf{Crim}} & {\textbf{Comm}} & {\textbf{Public}} & {\textbf{Lawyer}} &{\textbf{Avg}} 
 \\
 \hline
 Gem-2.5 & {\textbf{8.53}} & \textbf{8.28} & \textbf{8.27} & {\textbf{8.28}} & {\textbf{8.62}} & {\textbf{8.38}} \\
 GPT-4.1 & {8.44} & \textbf{8.28} & \textbf{8.27} & {8.14} & {8.48} & {8.35} \\
 Claude-3.7-S & {7.72} & 7.37 & 7.31 & {7.79} & {8.29} & 7.77 \\
 GPT-4.1-mini & {7.57} & 7.01 & 7.76 & {7.75} & {7.98} & 7.62 \\
 Gem-27B & 6.39 & \colorbox{highlightRed}{5.51} & \colorbox{highlightRed}{5.88} & 6.68 & 7.01 & 6.29 \\
 Krikri-8B & \colorbox{highlightRed}{5.95} & \colorbox{highlightRed}{5.84} & \colorbox{highlightRed}{5.79} & 6.78 & 6.74 & 6.20 \\
 \hline
 \end{tabular}
 }
 \caption{
  Comparison of model performance on different legal areas: `civil', `criminal', `commercial', `public', `lawyer' and the average score (`Avg').}
    \label{tab:fine-comparison}
\end{table}

\subsection{Chapter-based Legal Context}
\label{sec:context-analysis}

\begin{table*}[t]
\centering
\begin{tabular}{l@{\hspace{0.6em}}|c@{\hspace{0.6em}}c@{\hspace{0.6em}}c:c@{\hspace{0.6em}}|c@{\hspace{0.6em}}c@{\hspace{0.6em}}c:c@{\hspace{0.6em}}|c@{\hspace{0.6em}}c@{\hspace{0.6em}}c:c}
\hline
\textbf{Model} & \multicolumn{4}{c|}{\textbf{no context}} & \multicolumn{4}{c|}{\textbf{chapter context}} & \multicolumn{4}{c}{\textbf{oracle context}} \\
 & f & c & a & \textbf{avg} & f & c & a & \textbf{avg} & f & c & a & \textbf{avg} \\
\hline
{\footnotesize Gemini-2.5-Flash }& {\footnotesize\textbf{8.2}} & {\footnotesize\textbf{6.1}} & {\footnotesize\textbf{6.8}} & {\footnotesize\textbf{7.02}} & {\footnotesize 8.6} & {\footnotesize\textbf{8.2}} & {\footnotesize\textbf{8.4}} & {\footnotesize\textbf{8.38}} & {\footnotesize\textbf{8.9}} & {\footnotesize 8.8} & {\footnotesize 8.5} & {\footnotesize\textbf{8.73}} \\
{\footnotesize GPT-4.1 }& {\footnotesize 7.8} & {\footnotesize \colorbox{highlightRed}{5.8}} & {\footnotesize 6.4} & {\footnotesize 6.63} & {\footnotesize\textbf{8.7}} & {\footnotesize 8.1} & {\footnotesize 8.3} & {\footnotesize 8.35} & {\footnotesize 8.7} & {\footnotesize\textbf{8.9}} & {\footnotesize\textbf{8.6}} & {\footnotesize\textbf{8.73}} \\
{\footnotesize GPT-4.1-mini } & {\footnotesize 7.5} & {\footnotesize \colorbox{highlightRed}{4.4}} & {\footnotesize \colorbox{highlightRed}{5.4}} & {\footnotesize \colorbox{highlightRed}{5.73}} & {\footnotesize 8.3} & {\footnotesize 7.2} & {\footnotesize 7.4} & {\footnotesize 7.62} & {\footnotesize 8.7} & {\footnotesize 8.5} & {\footnotesize 8.1} & {\footnotesize 8.44} \\
{\footnotesize Gemini-2.0-Flash } & {\footnotesize 7.7} & {\footnotesize \colorbox{highlightRed}{5.0}} & {\footnotesize \colorbox{highlightRed}{5.5}} & {\footnotesize 6.05} & {\footnotesize 8.3} & {\footnotesize 7.1} & {\footnotesize 7.1} & {\footnotesize 7.51} & {\footnotesize 8.6} & {\footnotesize 8.2} & {\footnotesize 7.8} & {\footnotesize 8.19} \\
{\footnotesize Gemma-3-27B } & {\footnotesize 7.1} & {\footnotesize \colorbox{highlightRed}{4.1}} & {\footnotesize \colorbox{highlightRed}{4.9}} & {\footnotesize \colorbox{highlightRed}{5.37}} & {\footnotesize 7.5} & {\footnotesize \colorbox{highlightRed}{5.6}} & {\footnotesize \colorbox{highlightRed}{5.8}} & {\footnotesize 6.29} & {\footnotesize 8.5} & {\footnotesize 8.0} & {\footnotesize 7.7} & {\footnotesize 8.08} \\
\hline
\end{tabular}
\caption{Comparison with different context settings. Detailed scores for facts (f), cited articles (c), analysis (a) and mean/average (avg) are provided. The no-context setting makes the task much more challenging (most models barely get a passing score, which is above 6.00). The oracle context setting allows for higher scores, because the models make fewer mistakes in the cited articles score.}
\label{tab:no-context-extended}
\end{table*}

To investigate the impact of legal context, we experimented with three settings, with results presented in Table~\ref{tab:no-context-extended}. The ``no-context'' setting provided no legal corpus, forcing models to rely solely on their parametric knowledge. The ``chapter'' setting, our primary setup, provided relevant articles alongside distractors, as described in Section~\ref{sec:relevant-context}. The ``oracle'' setting provided only the exact correct articles. In the no-context setting, all models except Gemini-2.5-Flash performed poorly, underscoring their limited internal knowledge of Greek legislation. Conversely, the oracle setting enabled every model to achieve excellent scores, establishing that access to the correct legal statutes is the most critical factor for success in this task.

As expected, providing more accurate legal context—progressing from the \emph{no-context} to the \emph{chapters} and \emph{oracle} settings—dramatically improved the \emph{cited articles} score for all models. The \emph{facts} and \emph{analysis} scores also increased slightly, showing that access to the right articles improves the model's reasoning.

Notably, Gemini-2.0's initial lead over GPT-4.1-mini in the no-context setting (6.05 vs. 5.73) was overturned in the chapter (7.62 vs. 7.51) and oracle (8.19 vs 8.44) settings. This divergence demonstrates that strong parametric knowledge and the ability to process long context are separate skills, as shown in \citep{liu-etal-2024-lost}. This highlights the necessity of evaluating both abilities independently.

\section{Manual Error Analysis}
\label{sec:analysis}


To supplement our quantitative metrics, we conducted a manual error analysis to identify common failure modes. We analyzed the 10 worst-scoring responses from five key models. Human experts, guided by the scores and explanations of the LLM-judge, categorized the errors to reveal the most significant weaknesses. We focused on critical errors that led to an incorrect answer. The types of errors are the following:

\paragraph{Facts:} Errors include \textbf{hallucinations} (`Hall.'), inventing facts not present in the provided context, and \textbf{inaccuracies} (`Inac.'), misinterpreting or omitting key facts from the context.

\paragraph{Cited articles:} Errors include \textbf{inapplicable articles} (`Inap.'), citing legal articles not relevant to the case, and \textbf{missing articles} (`Missing'), failing to identify essential articles for the analysis.

\paragraph{Analysis:} Errors include an \textbf{invalid argument} (`Argum.'), a flawed argument from misunderstood principles or fallacious reasoning, and a \textbf{generic answer} (`Generic'), providing a general statement instead of an analysis tailored to the case facts.

Our analysis, while based on a small sample, highlights three primary areas of weakness: citing legal articles, understanding case facts, and constructing valid arguments. The most frequent errors related to failing to cite the correct legal \emph{articles} (23). This supports our finding (§\ref{sec:context-analysis}) that correct article retrieval is essential. \emph{Factual} (20) and \emph{analysis} (17) errors were also prevalent across all models. Although rare, critical fact hallucinations (3) appeared even in top-performing models, highlighting the risks of deploying these systems without human oversight.

\begin{table}[t!]
\resizebox{\columnwidth}{!}{%
\centering
\begin{tabular}{l cc cc cc}
\toprule
\textbf{Model} & \multicolumn{2}{c}{\textbf{Facts}} & \multicolumn{2}{c}{\textbf{Articles}} & \multicolumn{2}{c}{\textbf{Analysis}} \\
& \makecell{\small {Hall.}} & \makecell{\small Inac.} & \makecell{\small{Inap.}} & \makecell{\small Missed} & \makecell{ \small{Argum.}} & \makecell{\small Generic} \\
\midrule
Gemini-2.5-F & \textbf{1 }& 3 & 1 & 0 & 3 & \textbf{1} \\
GPT-4.1 & \textbf{1} & 2 & \textbf{3} & 5 & 2 & \textbf{1} \\
Claude-3.7-S & 0 & 3 & 0 & 3 & \textbf{5} & 0 \\
GPT-4.1-mini & 0 & 4 & 1 & 3 & 4 & 0 \\
Gemma-3-27B &\textbf{1} & \textbf{5} & 0 & \textbf{7} & 1 & 0 \\
\midrule
\textbf{Total} & \multicolumn{2}{c}{20} & \multicolumn{2}{c}{23} & \multicolumn{2}{c}{17} \\
\bottomrule
\end{tabular}
}
\label{tab:mistake_frequency_grouped}
\caption{Manual analysis of error types across different models. Errors are grouped into three main categories: Facts, Articles, and Analysis; and two subcategories for each one: hallucinations (`Hall.') and inaccuracies (`Inac.') for \emph{facts}, inapplicable (`Inap.') and missed (`Missed') \emph{articles}, and invalid argument (`Argum.') and generic answer (`Generic') for the \emph{analysis}.}
\end{table}

A key differentiator between human and model performance was the inability of LLMs to grasp subtle, legally significant context. For instance, all LLMs failed to infer that the phrase ``they had differences'' could indicate a motive for premeditation, which was obvious to human experts. This failure to interpret nuanced language was a common weakness, showing that significant improvement is needed for reliable use in real-world applications.

\section{Related Work}
\label{sec:related}


\paragraph{Legal domain:}
In the legal domain, LexGLUE \citep{chalkidis-etal-2022-lexglue} and LEXTREME \citep{niklaus-etal-2023-lextreme} are established benchmarks for legal classification tasks.
LegalBench \citep{legalbench} is the standard for evaluating LLMs on legal reasoning via multiple-choice questions. 
More closely related to our work, task 5.4 of LexEval~\citep{LexEval} is based on Chinese Bar exams, but, unlike our approach, it does not provide citations or use LLM-as-a-judge, instead evaluating with the less reliable, overlap-based ROUGE metric \citep{cohan-goharian-2016-revisiting}. Similarly, LLeQA \citep{lleqa} contains everyday legal questions in French, but evaluation is based on the METEOR metric without measuring its correlation with human experts. CaseGen~\citep{li2025casegenbenchmarkmultistagelegal} assesses document drafting and legal judgment generation in Chinese using the LLM-as-a-judge approach. 

Two concurrent works, OAB-Bench \citep{pires2025automatic} and LEXam \citep{fan2025lexam}, are most similar to GreekBarBench as they also provide open-ended legal questions and rubrics for LLM-judges. However, both have notable limitations. OAB-Bench, based on the Brazilian Bar exams, uses official guidelines as rubrics, but their complexity necessitates a costly model (OpenAI-o1), resulting in an evaluation cost of approximately \$50 per LLM. LEXam is based on English and German law school questions, which, unlike Bar exams, are targeted at less experienced students rather than graduates. Crucially, while both projects evaluate the quality of LLM-judges, neither provides these manual evaluations as a publicly available, standalone benchmark, preventing the community from using them to assess new judges in the future.

\paragraph{LLM-as-a-judge:}
LLM-as-a-judge was introduced by \citet{llmJudge}, who meta-evaluated its performance against human preferences for multi-turn chat assistant dialogues. A comprehensive overview of LLM-as-a-judge and meta-evaluation resources can be found in the survey by \citet{gu2024surveyllmasajudge}. Taking this concept further, JudgeBench \citep{judgeBench} introduced a general-purpose benchmark specifically for the meta-evaluation of LLM-judges. In line with our approach, other studies similarly develop separate benchmarks to meta-evaluate judges on specific tasks \citep{starace2025paperbenchevaluatingaisability,niklaus2025swiltrabenchswisslegaltranslation}.



\paragraph{Evaluation Rubrics:}
Legal research has for long focused on creating rubrics for consistent (human) evaluation of legal writing \citep{ClarkDeSanctis2013}. The Brazilian Bar exams have made their rubrics for human evaluation available, so the aforementioned OAB-Bench \citep{pires2025automatic} provides them to their LLM-judges. Their rubrics consist of a manually annotated ground truth answer with comments and a table with score distributions for each element of the answer. A proprietary benchmark, BigLawBench\footnote{\url{https://www.harvey.ai/blog/introducing-biglaw-bench}}, describes a scoring system that uses two dimensions: the `source' and `answer' scores, which are analogous to our \emph{Cited Articles} and \emph{Analysis}. They rely on detailed instructions per question that specify explicitly the attributes that would contribute positively and negatively to the final score of candidate answers. Constructing from scratch either of these approaches is prohibitively expensive, in contrast to our simple, span-based rubrics that only require minimal annotation effort. 


\paragraph{Greek NLP:}
Important Natural Language Processing resources for the Greek language include classification models \citep{greekBert, GreekLegalRoBERTa}, alongside more recent LLMs pretrained on Greek like Meltemi\footnote{Meltemi was excluded from our experiments, because of its relatively small context length of 8 billion tokens.} \citep{voukoutis2024meltemiopenlargelanguage} and Llama-Krikri-8B \citep{roussis2025krikriadvancingopenlarge}, which we tested in our experiments (\S~\ref{sec:leaderboard}). Existing Greek legal datasets cover only classification and summarization tasks \citep{angelidis2018named, papaloukas-etal-2021-multi, greeklegalsum}. Although Greek LLM benchmarks exist for other domains, such as finance \citep{peng2025plutusbenchmarkinglargelanguage} and medicine \citep{voukoutis2024meltemiopenlargelanguage}, the legal domain currently lacks one.

\section{Conclusions}
In this work, we introduced \textsc{GreekBarBench}, a benchmark evaluating LLMs on legal questions requiring citations to statutory articles and case facts. To enable robust and scalable assessment, we developed a comprehensive evaluation framework based on LLM-as-a-judge, which we validated against human experts using our GBB-JME meta-evaluation benchmark. Our results show that custom span-based rubrics significantly improve judge alignment. The extensive evaluation of 13 LLMs on \textsc{GreekBarBench} revealed that Gemini-2.5-Flash and GPT-4.1 achieved the best performance. Our primary finding is that success on this task is critically dependent on the quality of the provided legal context and the ability to identify the correct statutory articles within it.






\section*{Limitations}

Our benchmark, \textsc{GreekBarBench}, assumes the availability of the relevant legal code chapters for the \emph{Relevant Legal Context} component (\S~\ref{sec:relevant-context}). We did not evaluate the performance of retrieval models on this task, which is a critical step in real-world legal applications and could pose a significant challenge not addressed by our current setup.

A notable limitation is the cost associated with evaluating models using our framework due to the primary LLM-judge being a proprietary model (GPT-4.1-mini). To mitigate this cost, we suggest utilizing Simple-Judge with the open-weight model Gemma-3-27B. While no currently available open-weight model achieves meta-evaluation performance (SPA scores on GBB-JME) on par with GPT-4.1-mini, our public release of the benchmark and meta-evaluation dataset will allow future research to test and use more accurate and cost-effective LLM-judges.

\section*{Ethical Considerations}

The development and application of legal NLP benchmarks carry significant ethical implications and potential societal impact, particularly concerning fairness, access to justice, and responsible automation \citep{tsarapatsanis-aletras-2021-ethical}. Therefore, careful consideration of their design and potential uses is essential.

Our research contributes to the development of tools that could potentially assist various types of users, including legal professionals (such as judges and lawyers), students, and individuals seeking to understand legal concepts. It is crucial to emphasize that performance on this benchmark, or any similar research benchmark, should never be considered sufficient justification for deploying automated systems that substitute human experts. We strongly caution against the uncritical reliance on models evaluated solely on benchmark performance for automating legal tasks, making legal decisions, or providing legal advice. 


Despite our efforts to make \textsc{GreekBarBench} realistic, as a research benchmark, it overlooks two critical aspects for the safe and reliable deployment of legal AI applications in practice:

\begin{itemize}
    \item \textbf{Data Realism}: Real-world legal problems are far more complex and nuanced than the structured, often simplified scenarios found in exam questions \citep{medvedeva-mcbride-2023-legal}. They often demand significant legal interpretation, ethical judgment and persuasion, particularly when the law does not provide an explicit answer for a given situation.

    \item \textbf{Safety:} 
    Real-world applications must ensure that the AI system handles adversarial attacks effectively. Issues like guiding the decisions of the LLMs with malicious prompting (e.g., jailbreaking), and providing confident, incorrect information when asked legally unanswerable queries are unacceptable (see discussions on AI safety principles \footnote{\url{https://www.anthropic.com/news/core-views-on-ai-safety}}).

\end{itemize}

Furthermore, the primary ethical purpose of this work is not to provide a system ready for deployment, but to advance the state of legal NLP evaluation itself. By developing a benchmark that requires free-text generation, incorporates a multi-dimensional scoring system, and uses LLM-judges with explicit evaluation criteria, we aim to encourage the development of more transparent and explainable legal AI models. These features provide greater insight into how models arrive at their answers, moving beyond simple classification or multiple-choice and offering components of explainability which are crucial for gaining trust in AI applications \citep{medvedeva-mcbride-2023-legal}.

As already mentioned (\S~\ref{sec:exams}), the authors of the solutions of the exam papers have given approval for the public reproduction of this work, with respect to the original and strictly for academic research use. Our ground truth answers are based on the year that each exam was published. This means that if the relevant laws changed in the meantime, the solutions are no longer valid. All cases in the Greek Bar exams are fictional, created solely for educational purposes, and bear no relation to real individuals or actual legal cases.

\section*{Acknowledgments}
We are grateful to our legal expert annotators, Nasia Makridou and Irene Vlachou, for their diligent work, expertise, and insightful discussions, which were invaluable to this project.

This work was partially supported by project MIS 5154714 of the National Recovery and Resilience Plan Greece 2.0 funded by the European Union under the NextGenerationEU Program. AWS resources were provided by the National Infrastructures for Research and Technology (GRNET), with support from the EU Recovery and Resilience Facility.



\bibliography{anthology,custom}

\appendix
\break

\section{LLMs-as-judges variants}
\label{sec:llms-as-judges}

\begin{table}[ht]
\centering
\resizebox{\columnwidth}{!}{%
\begin{tabular}{lccc}
\toprule
Model & GPT-4.1-mini & Gemma-3-27B & Delta \\
 & Span & Simple &  \\
\midrule
Gemini-2.5-Flash & 8.38 & 8.53 & +0.15 \\
GPT-4.1 & 8.35 & 8.38 & +0.03 \\
OpenAI-o1 & 7.78 & 8.05 & +0.27 \\
Claude-3-7-Sonnet & 7.77 & \textbf{8.08} & +0.32 \\
GPT-4.1-mini & 7.62 & 7.87 & +0.25 \\
Gemini-2.0-flash & 7.51 & \textbf{7.88} & +0.37 \\
GPT-4o & 7.48 & 7.75 & +0.27 \\
Deepseek-R1 & 6.90 & 7.21 & +0.32 \\
Gemma-3-27b & 6.29 & 6.84 & +0.55 \\
Krikri-8B & 6.20 & 6.80 & +0.60 \\
Gemma-3-12b & 6.08 & 6.56 & +0.48 \\
GPT-4.1-nano & 5.45 & 5.77 & +0.31 \\
Gemma-3-4b & 4.42 & 5.03 & +0.61 \\
\bottomrule
\end{tabular}
}
\caption{Comparison of scores from two LLM-as-judge configurations: our primary judge (GPT-4.1-mini with \emph{Span-Judge}) versus an alternative (Gemma-3-27B with \emph{Simple-Judge}). The table is sorted by the GPT-4.1-mini scores. Scores from Gemma-3-27B are highlighted in \textbf{bold} where they alter the ranking of a model relative to its neighbors.}
\label{tab:other-judge}
\end{table}

Table~\ref{tab:other-judge} presents a comparative analysis to assess the robustness of our evaluation framework. We compare the scores from our primary judge, GPT-4.1-mini with the \emph{Span-Judge} prompt, against those from an alternative setup using Gemma-3-27B with a simpler \emph{Simple-Judge} prompt.

Our analysis reveals two key findings. First, the overall model rankings are highly consistent across both judges, as evidenced by a strong SPA agreement of 0.92. This indicates that our top-level conclusions about model performance are robust to the choice of the judge.

Second, despite the high rank correlation, we observe a significant and systematic scoring bias. Gemma-3-27B acts as a more lenient judge, assigning scores that are, on average, 0.35 points higher than GPT-4.1-mini. Crucially, this bias is not uniform across all models. As shown in the Delta column, the score inflation is most pronounced for lower-capability models (e.g., +0.61 for Gemma-3-4b and +0.60 for Krikri-8B) and is minimal for top-tier models. The exceptionally small delta for GPT-4.1 (+0.03) is particularly noteworthy. This suggests that the judges disagree on its proximity to the top-ranked Gemini-2.5-Flash. The low delta could indicate a potential "
``family bias,'' where the GPT-4.1-mini judge evaluates its sibling model more favorably, rating it closer to the top performer than the external Gemma judge does.

This non-uniform bias leads to minor but notable disagreements in the fine-grained ordering of similarly-performing models. For example, the Gemma judge awards a disproportionately higher score to Claude-3-7-Sonnet and Gemini-2.0-Flash, causing them to overtake their immediate competitors in the ranking, as highlighted in bold. This experiment underscores that while overall rankings are stable, the absolute scores and the precise ordering of closely-ranked models can be sensitive to the specific judge and prompting method used.



\section{Contamination}
\begin{table*}[ht]
\centering
\begin{tabular}{|l|c|c|c|c|}
\hline
\textbf{Model} & \textbf{Knowledge cutoff} & \textbf{GreekBarBench} & \textbf{Semi-private} & \textbf{Delta} \\ \hline
Gemini-2.5-Flash & \textbf{2025-01-01} & 8.40 & 8.04 & -- 0.36 \\ \hline
GPT-4.1 & 2024-06-01 & 8.32 & 7.65 & -- 0.67 \\ \hline
OpenAI-o1 & 2023-09-01 & 7.77 & 7.51 & -- 0.26 \\ \hline
Claude-3.7-Sonnet & 2024-09-30 & 7.71 & 7.51 & -- 0.20 \\ \hline
GPT-4.1-mini & 2024-06-01 & 7.63 & 7.09 & -- 0.54 \\ \hline
Gemma-3-27B & 2024-08-30 & 6.33 & 5.51 &\textbf{-- 0.82} \\ \hline
Krikri-8B & \textbf{2025-02-09} & 6.24 & 5.56 & -- 0.68 \\ \hline
\multicolumn{5}{|l|}{\textit{Semi-private set cutoff: 2024-10-03}} \\ \hline
\end{tabular}
\caption{We report the overall performance of LLMs on the original GreekBarBench and on the semi-private set. Delta is the performance difference between the two sets. The models highlighted in bold could potentially have contaminated data and still trick the test, because their cutoff date is after the semi-private set's date of  
publication.}
\label{tab:contamination}
\end{table*}

Table~\ref{tab:contamination} compares the performance of several LLMs on the public GreekBarBench and its `semi-private' version. A significant drop in performance on the `semi-private' set, indicated by a large negative delta, would suggest contamination. The results show no clear evidence of such contamination. The observed performance drops are minor for all models (under a 1.0-point delta), suggesting the `semi-private' set is simply more difficult. The largest deltas are also associated with the lowest-performing models (Gemma-3-27B and Krikri-8B), which contradicts the expected score inflation.

The exceptions to this analysis are Gemini-2.5-Flash and Krikri-8B, whose late data cutoffs allow for potential data exposure to remain undetected. However, the concern for Krikri-8B is mitigated by its very low performance, which is inconsistent with contamination, as mentioned before. Consequently, only the results for Gemini-2.5-Flash are inconclusive. Our final assessment is that most models appear uncontaminated, while the status of Gemini-2.5-Flash cannot be determined.




\section{Annotator Instructions}
\label{app:annotator-instructions}
In this section we present the instructions given to the two legal expert annotators. The annotators possessed prior experience with the evaluation task, having previously taken the exams themselves. This existing expertise allowed for concise instructions. For the general evaluation of LLM-generated answers for the manual evaluation (\S~\ref{sec:manual}), the instruction (translated to English) was simply to:
\begin{quote}
``Evaluate the candidate answers on each scoring dimension (\emph{Facts}, \emph{Cited Articles}, and \emph{Analysis}).''
\end{quote}

For the creation of text spans for Span-Judge (\S~\ref{sec:span_llm_judge}), annotators were instructed to:
\begin{quote}
``Highlight the text-spans that correspond to each scoring dimension (\emph{Facts}, \emph{Cited Articles}, and \emph{Analysis}). Highlight the most important subsets of these spans with the label \emph{important}.''
\end{quote}

\section{Complete Prompts}
\label{app:prompts}
\begin{figure*}[ht]
    \centering
    \includegraphics[
        width=\linewidth,
        trim=0pt 740pt 0pt 0pt,
        clip
    ]{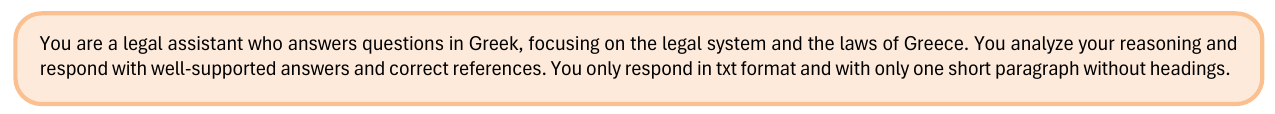}
    \caption{System prompt for generation given to candidate LLMs.}
    \label{app:gen-system-prompt}
\end{figure*}

\begin{figure*}[ht]
    \centering
    \includegraphics[
        width=\linewidth,
        trim=0pt 700pt 0pt 0pt,
        clip
    ]{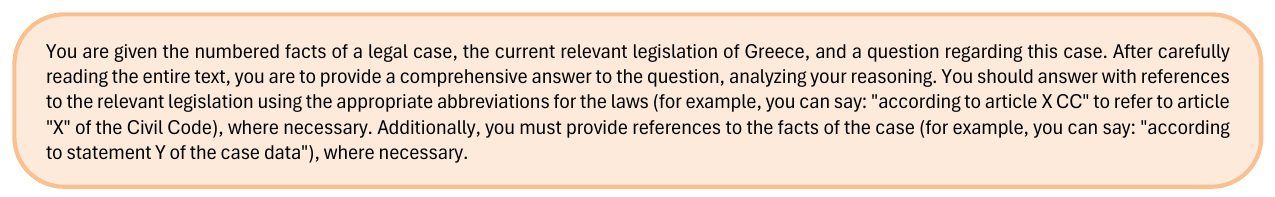}
    \caption{User prompt for generation given to candidate LLMs.}
    \label{app:gen-user-prompt}
\end{figure*}

\begin{figure*}[ht]
    \centering
    \includegraphics[
        width=\linewidth,
        trim=0pt 450pt 0pt 0pt,
        clip
    ]{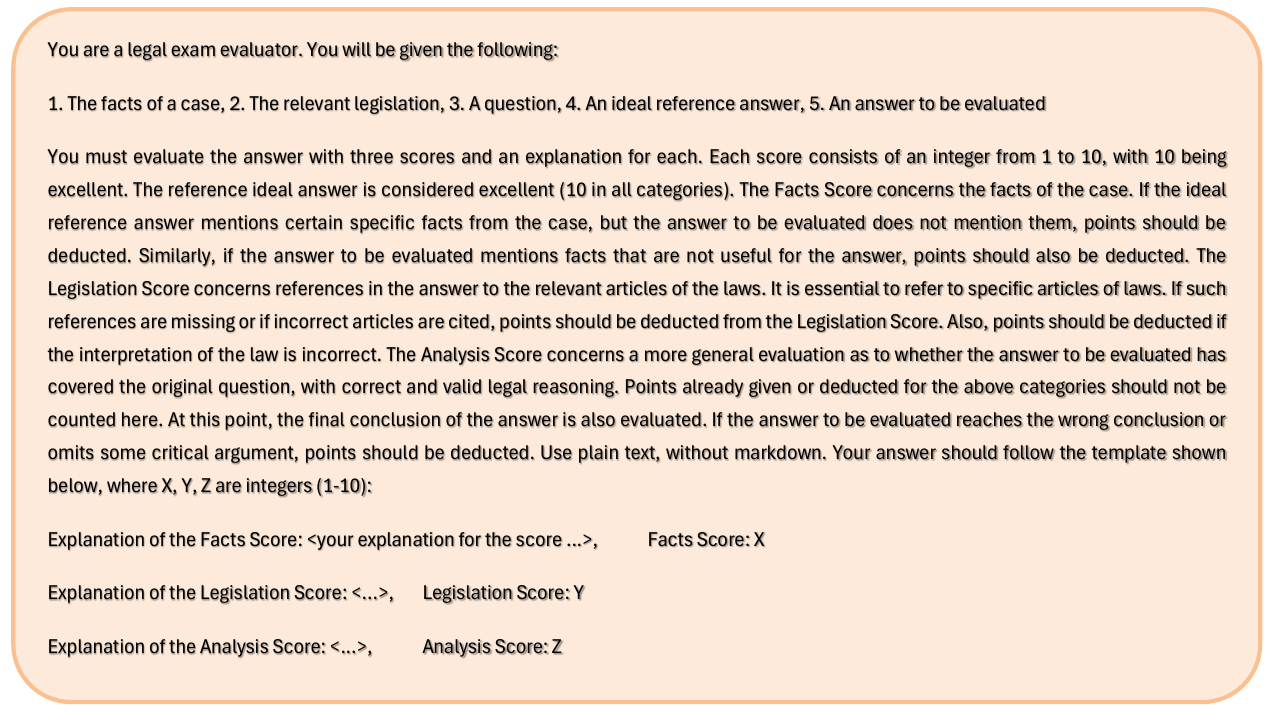}
    \caption{System prompt given to Simple-Judge LLM-judges.}
    \label{app:simple-judge-prompt}
\end{figure*}

\begin{figure*}[ht]
    \centering
    \includegraphics[
        width=\linewidth,
        trim=0pt 320pt 0pt 0pt,
        clip
    ]{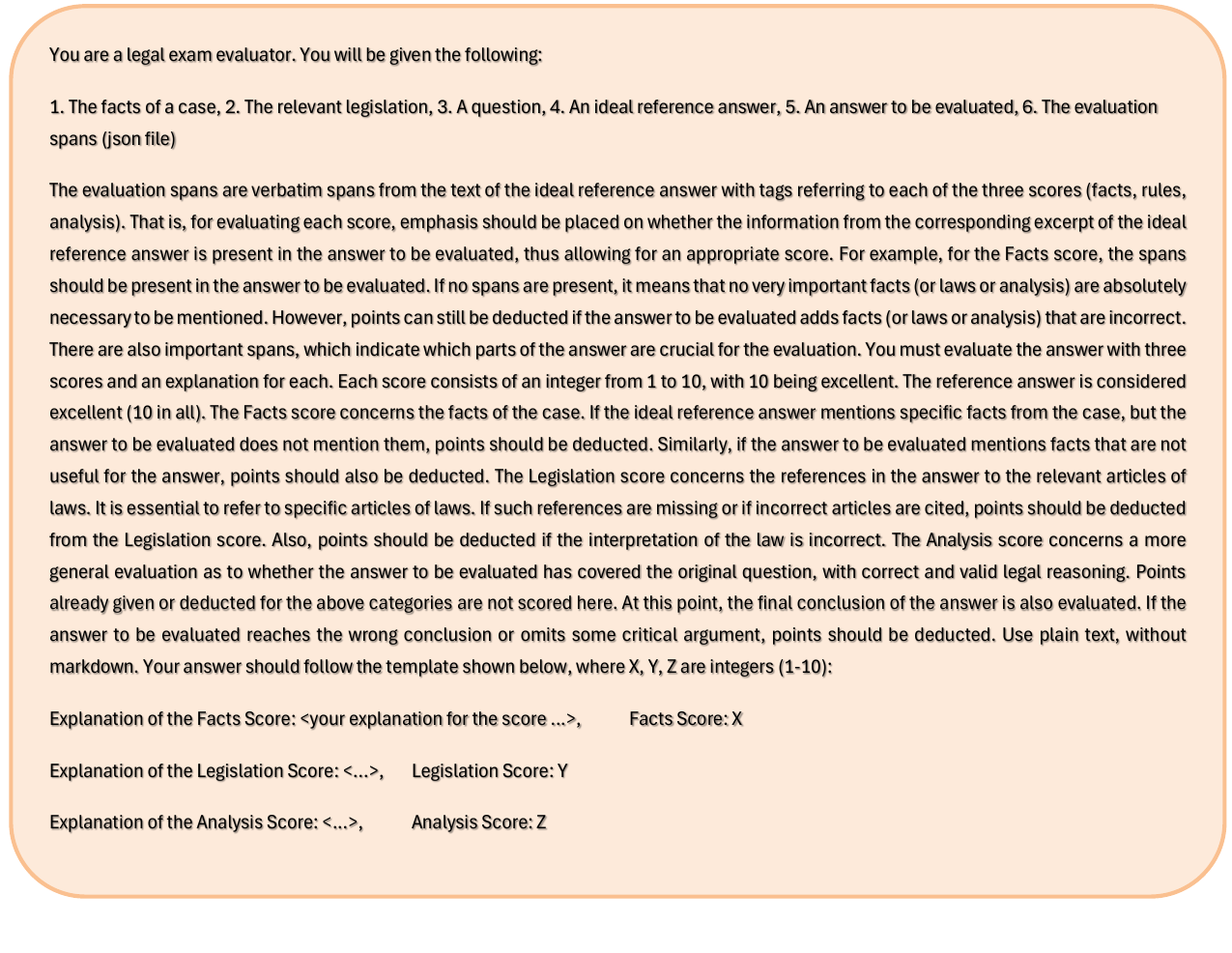}
    \caption{System prompt given to Span-Judge LLM-judges.}
    \label{app:span-judge-prompt}
\end{figure*}

In this section we present the complete system (Fig.\ref{app:gen-system-prompt}) and user (Fig.\ref{app:gen-user-prompt}) prompts given to candidate LLMs for generation of answers, as well as the complete system prompts given to LLM-judges for the Simple- (Fig.\ref{app:simple-judge-prompt}) and Span-Judge (Fig.\ref{app:span-judge-prompt}).

\section{Complete Dataset Example}
\label{app:example}
\begin{figure*}[ht]
    \centering
    \includegraphics[
        width=\linewidth,
        trim=20pt 100pt 20pt 30pt,
        clip
    ]{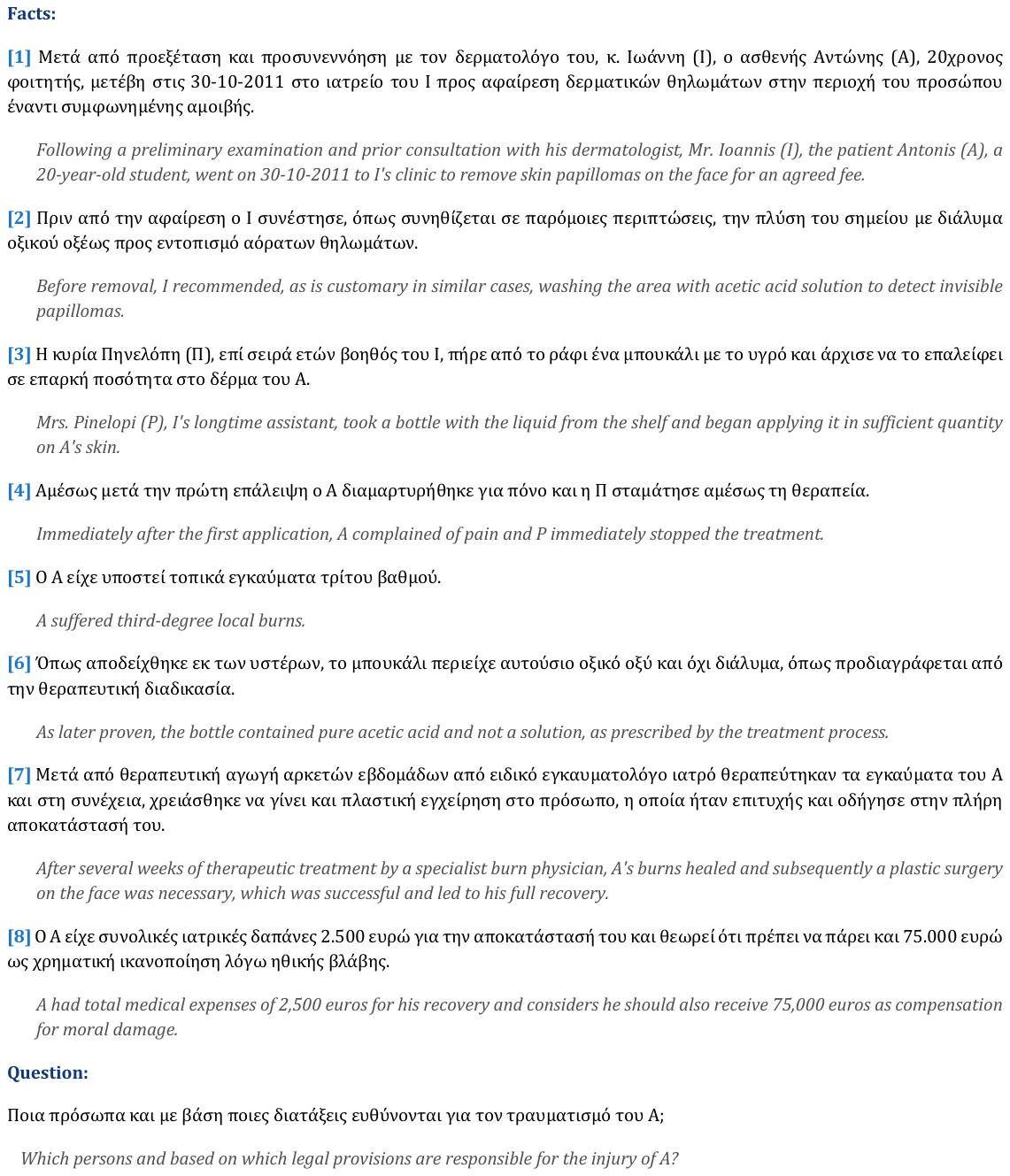}
    \caption{Complete \emph{Facts} and \emph{Question} (original and below translated in English), as given in to the candidate LLMs, for the example in Table~\ref{tab:example}.}
    \label{app:example-facts}
\end{figure*}

\begin{figure}[ht]
    \centering
    \includegraphics[
        width=\linewidth,
        trim=130pt 320pt 210pt 30pt,
        clip
    ]{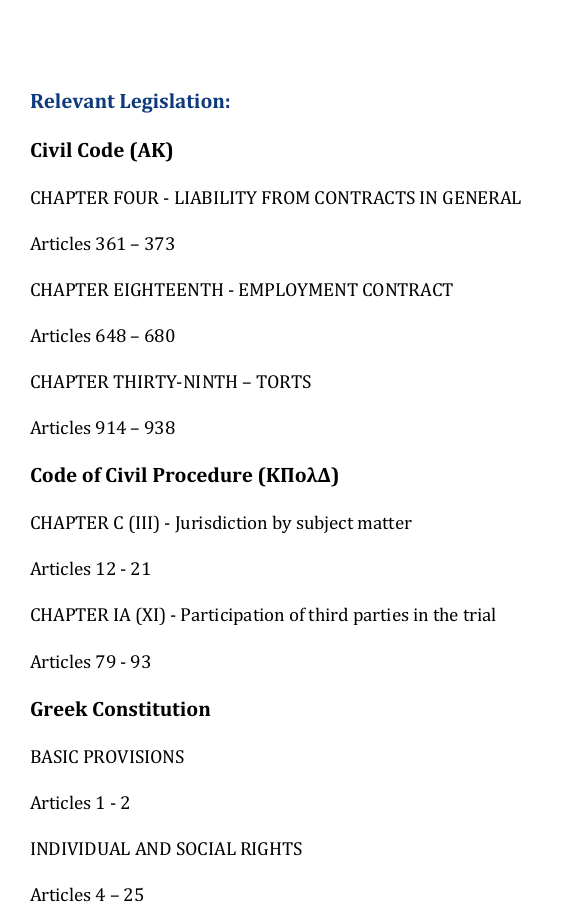}
    \caption{The \emph{Chapters} of the \emph{Relevant Legislation} context given to candidate LLMs, for the example in Table~\ref{tab:example}. The content of the articles is not shown for brevity.}
    \label{app:legislation-main}
\end{figure}

\begin{figure}[ht]
    \centering
    \includegraphics[
        width=\linewidth,
        trim=180pt 100pt 180pt 30pt,
        clip
    ]{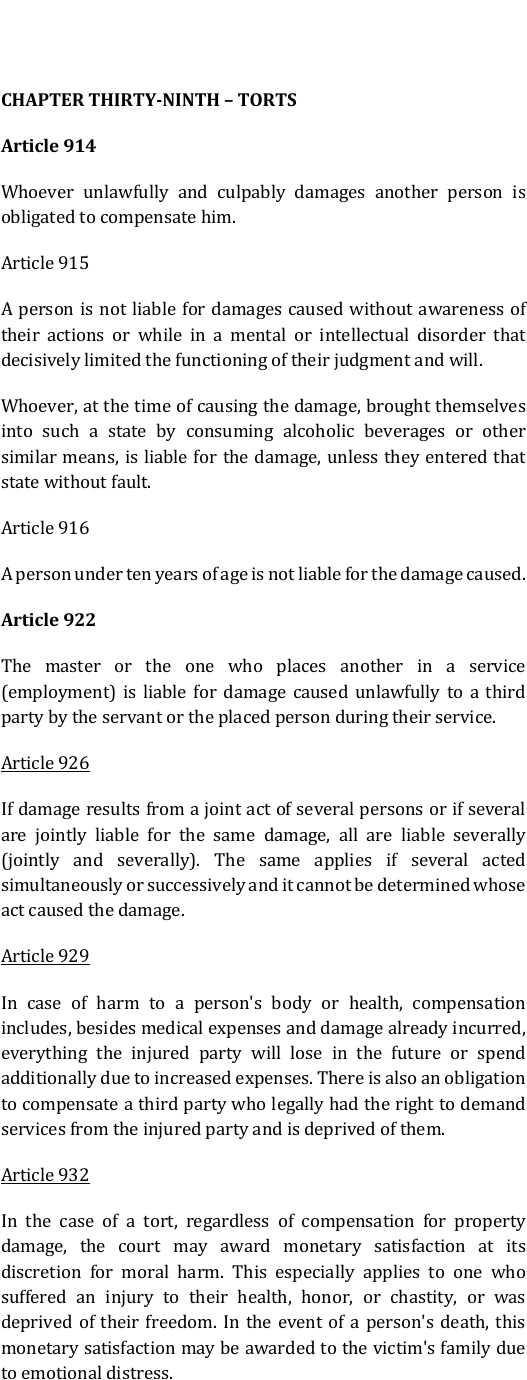}
    \caption{Chapter Thirty-Ninth (`TORTS') from the Civil Code, which is part of the \emph{Relevant Legislation} context given to candidate LLMs, for the example in Table~\ref{tab:example}. The \emph{gold} cited articles are marked in bold and the articles cited by Gemini-2.5-Flash(Figure~\ref{app:gemini-answer}) are underlined.}
    \label{app:legislation-chapter}
\end{figure}

\begin{figure*}[ht]
    \centering
    \includegraphics[
        width=\linewidth,
        trim=20pt 120pt 20pt 30pt,
        clip
    ]{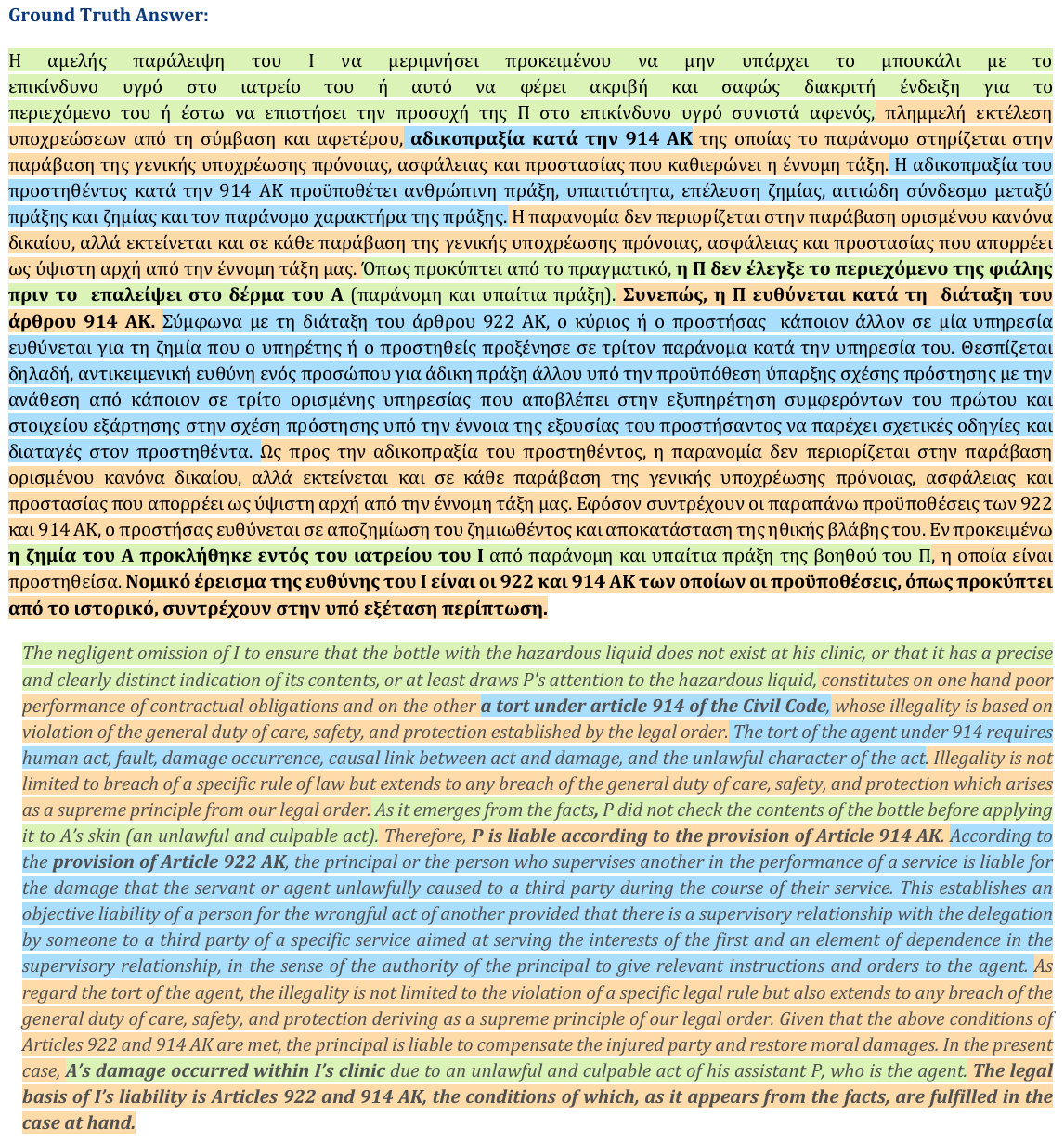}
    \caption{Ground truth answer by the legal expert. Text spans are highlighted in colors (green for \emph{Facts}, blue for \emph{Cited Articles} and orange for \emph{Analysis}).}
    \label{app:ground-truth}
\end{figure*}

\begin{figure*}[t]
    \centering
    \includegraphics[
        width=\linewidth,
        trim=20pt 460pt 20pt 30pt,
        clip
    ]{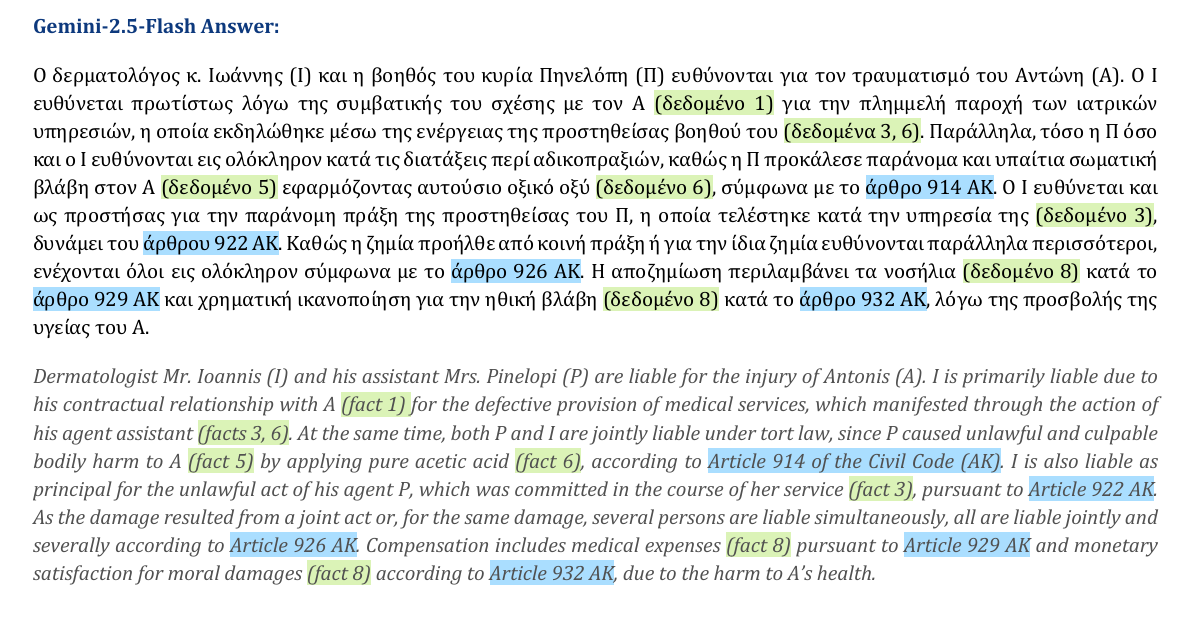}
    \caption{The answer of Gemini-2.5-Flash for the example in Table~\ref{tab:example}. Citations are highlighted in color (green for \emph{Facts} and blue for \emph{Cited Articles}).}
    \label{app:gemini-answer}
\end{figure*}

\begin{figure*}[ht]
    \centering
    \includegraphics[
        width=\linewidth,
        trim=20pt 350pt 20pt 30pt,
        clip
    ]{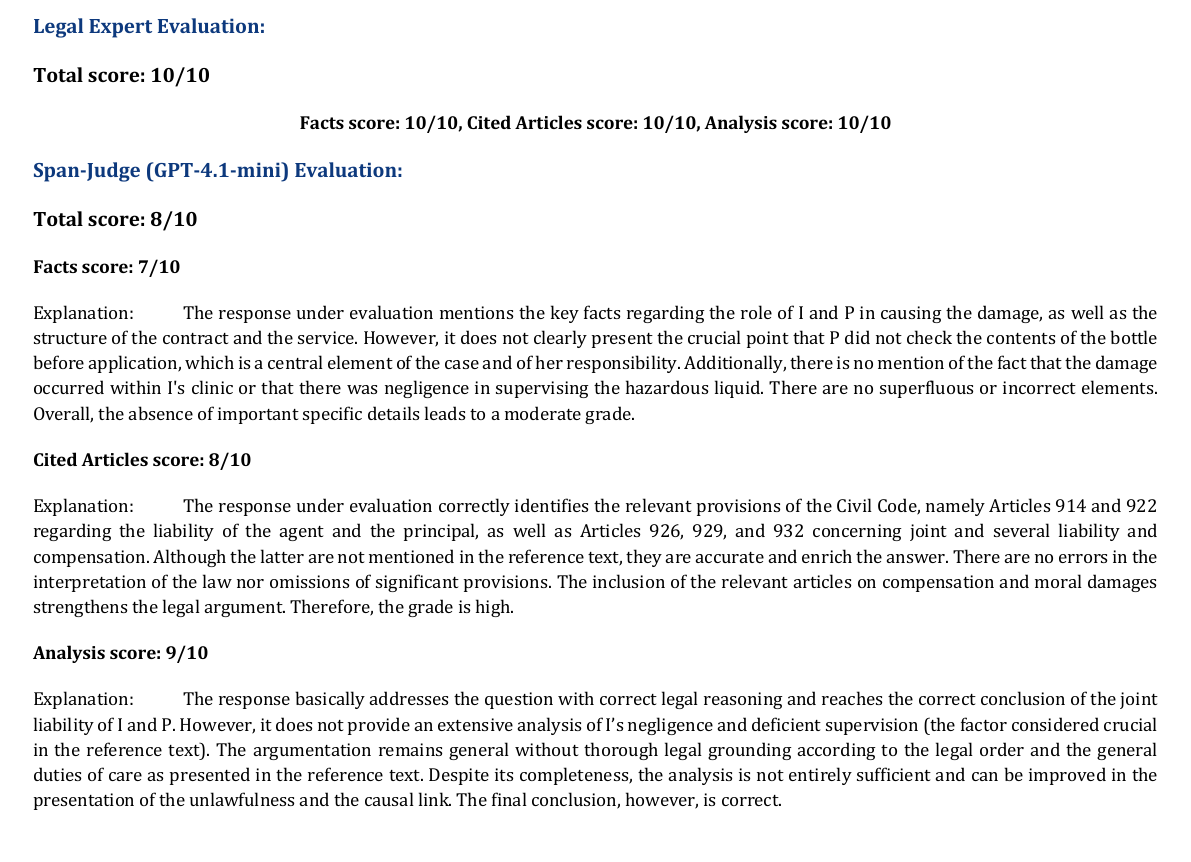}
    \caption{Evaluation results for Gemini's answer by Legal Experts and the LLM-Judge (GPT-4.1-mini Span-Judge). The response is perfect according to the legal experts. The LLM-judge is more strict and gives an 8/10 total score.}
    \label{app:evals}
\end{figure*}

In this section we present the complete version of the example that we presented in Table~\ref{tab:example}. We show the complete \emph{Facts} and \emph{Question} (Fig.~\ref{app:example-facts}), the \emph{Relevant Legal Context} (Figures~\ref{app:legislation-main} and \ref{app:legislation-chapter}), the complete \emph{Ground Truth Answer} (Fig.~\ref{app:ground-truth}), the candidate answer by Gemini-2.5-Flash (Fig.~\ref{app:gemini-answer}) and \emph{evaluations} of the candidate answer by the legal experts and the LLM-judge (Fig.\ref{app:evals}).

\end{document}